# Extending Factor Graphs so as to Unify Directed and Undirected Graphical Models

Brendan J. Frey
University of Toronto
http://www.psi.toronto.edu

## ABSTRACT

*The two most popular types of graphical model are Bayesian networks (BNs) and Markov random fields (MRFs). These types of model offer complementary properties in model construction, expressing conditional independencies, expressing arbitrary factorizations of joint distributions, and formulating message-passing inference algorithms. We show how the notation and semantics of factor graphs (a relatively new type of graphical model) can be extended so as to combine the strengths of BNs and MRFs. Every BN or MRF can be easily converted to a factor graph that expresses the same conditional independencies, expresses the same factorization of the joint distribution, and can be used for probabilistic inference through application of a single, simple message-passing algorithm. We describe a modified "Bayes-ball" algorithm for establishing conditional independence in factor graphs, and we show that factor graphs form a strict superset of BNs and MRFs. In particular, we give an example of a commonly-used model fragment, whose independencies cannot be represented in a BN or an MRF, but can be represented in a factor graph. For readers who use chain graphs, we describe a further extension of factor graphs that enables them to represent properties of chain graphs.*

## 1  INTRODUCTION

Graphical models play an important role in the theory, practice and communication of probability models and associated inference algorithms. Pictorial representations of graphical models are useful for efficiently communicating the conditional independence structure in a system of random variables, efficiently communicating the factorization of the joint distribution into functions on subsets of variables, and graphical user interfaces for constructing graphical models. Even for large scale systems that may contain millions of random variables, pictorial representations are useful for describing fragments of the model, model structures that occur repeatedly, and model structures on large groups of variables. Leaving pictures aside, graphical models are useful for deriving computational techniques for analyzing independence relationships, and deriving exact and approximate probabilistic inference algorithms.

The two most popular types of graphical model, BNs (BNs) (Pearl 1988) and Markov random fields (MRFs) (Kinderman and Snell 1980) offer complementary strengths. Whereas BNs are well-suited to construction using known causal relationships, MRFs are well-suited to construction using known Markov blankets, or non-causal relationships. Whereas BNs are well-suited to expressing *marginal independence* and *conditional dependence* ("explaining away"), MRFs are well-suited to expressing *marginal dependence* and *conditional independence*. Researchers choose to use one type or the other for a particular application, or, as is often the case, prefer to use one or the other in all of their work, potentially limiting the scope of their work.

BNs and MRFs also have common weaknesses. While bucket elimination (Dechter 1996) offers a single, concise algorithm for exact inference in both models, approximate probability propagation algorithms that ignore cycles have been recently used to achieve breakthroughs on difficult instances of NP-hard problems, including problems in error-correcting decoding (MacKay and Neal 1996; Frey and Kschischang 1996; McEliece 1997), phase unwrapping (Frey, Koetter and Petrovic 2002), and satisfiability (Mézard, Parisi and Zecchina 2002). Local message-passing inference algorithms are notationally cumbersome in BNs (Pearl 1988) and extra rules apply when the BN is a "polytree", a term that MRF researchers are reluctant to incorporate into their vocabularies. Local message-passing inference algorithms cannot be implemented directly on MRFs. Instead, the MRF must be converted to a graph on groups of replicated variables, which usually obscures many of the conditional inde-pendence relationships that BN researchers prefer to keep clear.

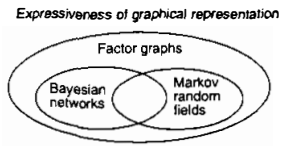

Figure 1: For any BN or MRF, there is a factor graph that expresses the same conditional independencies and the same factorization of the join distribution. Importantly, there are interesting and useful factor graphs that have properties that cannot be expressed by a BN or MRF.

Another common weakness of BNs and MRFs is that neither model can express an arbitrary factorization of the joint distribution. Often, the joint distribution over a set of variables is best described by the product of a large number of functions that overlap in their dependencies. A simple example is $P(x,y) = f(x,y)g(x,y)$, whose factorization cannot be expressed using a BN or an MRF on $x$ and $y$. Examples of real-world models of this sort that have recently taken center stage in the AI and information theory communities include low-density parity-check codes (Gallager 1963; MacKay and Neal 1996) and random satisfiability problems (Mézard, Parisi and Zecchina 2002). For these kinds of models, neither a BN nor an MRF on the same set of variables can be used to explicitly identify all of the functions. While this is not important for ascertaining conditional independence relationships, the ability to specify arbitrary factorizations is important when communicating models, and can have a tremendous impact on the efficiency and accuracy of inference algorithms.

Chain graphs (Lauritzen 1996) are graphical models that can have both directed and undirected edges. While they combine some of the conditional independence properties of BNs and MRFs, they are not well-suited to deriving message-passing algorithms, and like BNs and MRFs, they are not able to represent arbitrary factorizations of the joint distribution.

In (Frey et al. 1998; Kschischang, Frey and Loeliger 2001), we introduced *factor graphs* as a graphical model that can explicitly represent arbitrary factorizations of joint distributions, and are especially well-suited to generalized message-passing algorithms. Unlike BNs, MRFs and chain graphs, factor graphs explicitly identify terms in the joint distribution through the use of *function nodes*. This graphical notation leads to a single local message-passing algorithm that can be used to perform inference in both directed and undirected models. In fact, the explicit identification of functions enables message passing in factor graphs to be more efficient than belief propagation and the junction tree algorithm, in some cases.

In this paper, we show how factor graphs can be extended so as to unify the conditional independence properties of BNs and MRFs. Factor graphs of this new type can be constructed from knowledge of causal relationships, non-causal relationships, or both, and they can express all conditional independencies that can be expressed by BNs and MRFs. While every BN or MRF has an equally expressive factor graph, there are factor graphs whose properties cannot be expressed by a Bayesian network or an MRF.

In the next section, we define our extension of factor graphs and show how BNs and MRFs can be easily converted to factor graphs, without loss of semantics. We also describe simple, easily verifiable conditions that are sufficient for converting a factor graph to a BN or an MRF. Then, we describe the rule for ascertaining conditional independence in factor graphs – a rule which naturally follows from the rules for BNs and MRFs. We show that insofar as expressing conditional independencies and factorization, factor graphs subsume BNs and MRFs. For readers interested in chain graphs, we show how a further extension of factor graphs can be used to represent properties of chain graphs, in particular the chain graph $a \to c - d \leftarrow b$. We finish with some concluding remarks.

## 2 EXTENDING FACTOR GRAPHS

Graphical models describe how the joint distribution over a set of variables can be decomposed into a product of functions on subsets of variables.

**Original factor graphs.** In (Frey et al. 1998; Kschischang, Frey and Loeliger 2001), we define a factor graph for variables $x_1, \ldots, x_N$ and functions $g_1, \ldots, g_K$, to be a bipartite graph on a set of nodes corresponding to the variables, and a set of nodes corresponding to the functions. Each function depends on a subset of the variables, and the corresponding function node is connected to the nodes corresponding to to the subset of variables, i.e., if $g_k$ depends on $x_i$, there is an edge connecting $g_k$ and $x_i$. The joint distribution is given by

$$P(x_1, \ldots, x_N) = \prod_{k=1}^{K} g_k(x_{C_k}), \qquad (1)$$

where $C_k$ is the index set of the variables that are connected to function $g_k$, and $x_{C_k}$ denotes this set of variables. Like BNs, many factor graphs are automatically normalized, but like MRFs, some factor graphs require normalization. In the latter case, we include a constant $g_0 = 1/((\sum_{x_1,\ldots,x_N} \prod_{k=1}^{K} g_k(x_{C_k}))$ to ensure that $P(x_1, \ldots, x_N)$ is normalized.

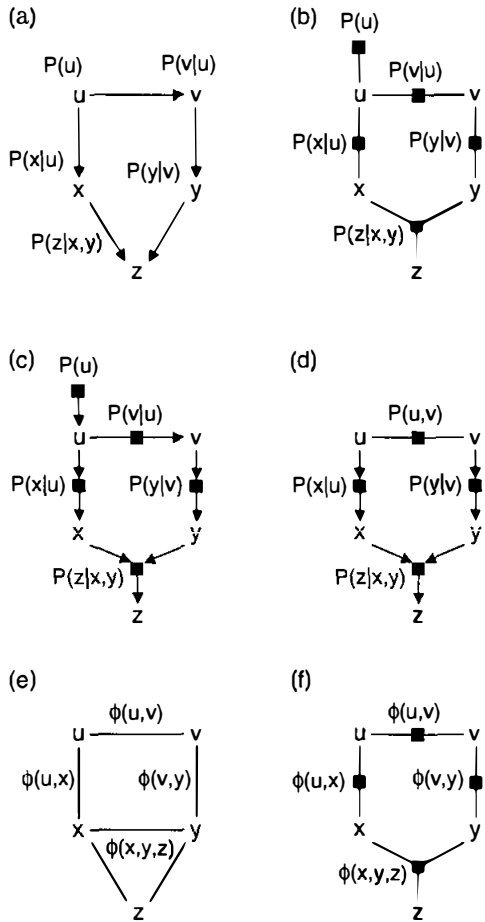

Figure 2: (a) A BN. (b) The factor graph obtained by creating one function node (solid black square) for each variable, and connecting each function node to the corresponding child and its parents. (c) A directed factor graph can be used to indicate conditional probabilities. (d) A hybrid construction, where $u$ and $v$ are modeled in an undirected fashion as $P(u,v)$ and $x$, $y$ and $z$ are modeled in a directed fashion. (e) A Markov random field (MRF). (f) The factor graph obtained by creating one function node for each maximal clique, and connecting the variables in each maximal clique to the corresponding function node.

Fig. 2a shows a BN that describes the following factorization:

$$P(u,v,x,y,z) = P(u)P(v|u)P(x|u)P(y|v)P(z|x,y).$$

This joint distribution is a product of 5 functions, so we can represent it using a factor graph with 5 function nodes. Fig. 2b shows a factor graph where the function nodes are solid black squares, and the corresponding functions are written beside each function node. Frequently, discs are used to indicate variables in a graphical model, but for visual clarity, we indicate each variable node with its corresponding random variable.

**Extension of factor graphs.** To represent properties of directed models, we allow factor graphs to have directed edges, which indicate conditional distributions. A simple example is shown in Fig. 2c. In parent-child relationships, the children are indicated by edges directed from functions to variables and the parents are indicated by edges directed from variables to functions. As with BNs, a factor graph must not have directed cycles. If every edge in the factor graph is directed, we refer to the model as a *directed factor graph*. The above extension has implications for local normalization requirements and the rule for ascertaining conditional independencies. So as to ease the reader into these implications, we describe some simple examples and the straight-forward correspondences between factor graphs and BNs and MRFs, before formalizing the normalization requirements and describing a Bayes-ball algorithm for ascertaining conditional independence.

Fig. 2c shows a directed factor graph for the BN in Fig. 2a. Notice that by looking at the graph alone, it is possible to identify the conditional distributions in Fig. 2c, whereas it is not possible to do so in Fig. 2b. In this example, each variable has just one incoming directed edge and each function has just one outgoing directed edge. However, we will see later that in general, each variable can have multiple "parent functions" and each function can have multiple "child variables".

Often, we would like to model some parts of a system in an undirected fashion (using noncausal knowledge), whereas we'd like to model other parts of a system in a directed fashion (using causal knowledge). An example is a model of a magnetic disk medium and the noisy read/write head which sometimes causes errors. An undirected Ising model may be appropriate for modeling deterioration of information on the magnetic disk, whereas a directed channel noise model may be appropriate for modeling the noise introduced by the read/write head. Fig. 2d shows a factor graph for the BN in Fig. 2a, where $u$ and $v$ are modeled in an undirected fashion as $P(u,v)$, while the rest of the variables are modeled in a directed fashion as before. Chain graphs can also represent hybrid constructions, but are less specific about factorization, as discussed below.

To contrast directed and semi-directed factor graphs with undirected factor graphs, consider the MRF in Fig. 2e. This MRF corresponds to the BN in Fig. 2a, and describes the factorization, $P(u,v,x,y,z) = \phi(u,v)\phi(u,x)\phi(v,y)\phi(x,y,z)$. This joint distribution is a product of 4 functions, so we can represent it using a factor graph with 4 function nodes. Fig. 2f shows the corresponding factor graph.

## 2.1 From BNs to Factor Graphs and Back

Every BN can be converted to a factor graph without loss of semantics, although the converse is not true. To convert a BN to a factor graph, imagine the edges in the BN are elastic bands. For each child in the BN the incoming directed edges are pinched together and "pinned" with a function node whose corresponding function is set equal to the conditional probability of the child given its parents. The edges connecting the parents and the function are directed toward the function. If a variable does not have any parents, a function node corresponding to the variable's marginal distribution is created and connected to the variable with a directed edge. See Fig. 2a and Fig. 2c for an example. To convert the factor graph back to a BN, simply "un-pin" each function node and set the conditional probability of the child given its parents equal to the function corresponding to the function node. The resulting BN is identical to the original BN.

The edge complexity of the factor graph obtained from a BN is similar to the complexity of the BN. If the BN has $E$ edges and $N$ variables, the factor graph will have $E + N$ edges.

Since the conversion process is reversible, all semantics associated with the BN carry over to the factor graph. In fact, when looking at the factor graph, it is easy to imagine the functions "unpinned" and directly apply the graph-based rule for ascertaining independencies in the BN (Pearl 1988).

However, although the factor graph for a BN retains all semantics of the BN, later in this paper we show that the converse is not true. There are directed factor graphs whose corresponding BNs are not as expressive. Later, we describe how to ascertain conditional independencies in factor graphs and find that there are directed factor graphs that can express independencies that cannot be expressed in BNs.

## 2.2 From MRFs to Factor Graphs and Back

As with BNs, every MRF can be converted to a factor graph without loss of semantics. To convert an MRF to a factor graph, one function node is created for each maximal clique and each function node is connected to the variables in the corresponding maximal clique. Each function is set to the potential function on the corresponding maximal clique of variables. See Fig. 2e and Fig. 2f for an example, where there are 4 maximal cliques. To convert the factor graph back to the original MRF, for each function node, a clique is created on the neighboring variables. These cliques are maximal, since the function nodes were created from maximal cliques. The potential for each maximal clique is set to the corresponding function in the factor graph.

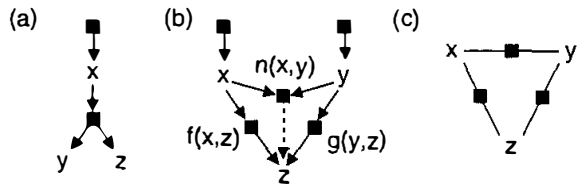

Figure 3: Factor graphs whose BNs and MRFs are less expressive. (a) A directed factor graph that distinguishes between the children in the conditional distribution $P(y, z|x)$. (b) A directed factor graph for a model where the conditional distribution $P(z|x, y)$ factorizes. (c) An undirected factor graph for a model with 3 unique potentials on 3 pairs of variables.

The edge complexity of the factor graph can be significantly lower than the edge complexity of the MRF. For an MRF with $E$ edges and $N$ variables, the factor graph will have at most $E + N$ edges, but may have $N$ times fewer edges. The latter case is exemplified by a fully-connected MRF on $N$ variables, which has order $N^2$ edges, whereas the factor graph has $N$ edges.

Applying the above two procedures to convert an MRF to a factor graph and then back to an MRF, always leaves the MRF unchanged. So, the semantics of MRFs carry over to factor graphs. For example, it is easy to show that the Markov blanket is given by a variable's second neighbors in the factor graph. In Fig. 2f, the Markov blanket of $y$ is $\{v, x, z\}$. Later, we show that the converse is not true – there are factor graphs whose corresponding MRFs are not as expressive.

## 3 BEYOND BNs AND MRFs

Fig. 3 shows examples of factor graphs that convey more structural information than any BN or MRF for the same distribution. The first example demonstrates how a factor graph can distinguish between multiple children in a conditional distribution. While a BN can "fake" this by created fully-connected subgraphs-graphs (e.g. on $x$, $y$ and $z$), the resulting network implies a directional dependence that may not exist. The second example shows how a factor graph can represent a conditional distribution of the form $P(z|x, y) = f(x, z)g(y, z)n(x, y)$, where $n(x, y) = 1/\sum_z f(x, z)g(y, z)$. The fact that $n(x, y)$ is a normalization function for $z$ may be indicated by a dashed edge from $n(x, y)$ to $z$, although this is not formally required. While the factor graph can represent this factorized form, a BN on $x$, $y$ and $z$ cannot. The third examples shows how an undirected factor graph can represent a distribution on 3 variables as a product of 3 unique potentials. The MRF for this distribution contains one maximal clique and so cannot indicate the existence of 3 unique potentials.

### 3.1 Local Normalization Condition

The product of all functions in a directed factor graph will give a normalized joint distribution, if the local normalization condition described below is satisfied. We describe various situations, before giving the general normalization condition.

If a function has exactly one outgoing edge that is connected to a variable that has exactly one incoming edge, the function must be normalized w.r.t. the variable. For example, the function connected to $x$, $y$ and $z$ in Fig. 2c must be normalized w.r.t. $z$. If a function has multiple outgoing edges that are connected to variables that each have exactly one incoming edge, the function must be normalized w.r.t. the set of variables. For example, the function connected to $x$, $y$ and $z$ in Fig. 3a must be normalized w.r.t. the pair $y, z$.

Unlike BNs, directed factor graphs can represent a conditional distribution as a product of functions. Such a situation is indicated by a variable that has multiple incoming directed edges, where the conditional distribution is given by the product of the functions that are connected to the variable by directed edges. However, to ensure normalization, there may be functions connecting all parents or subsets of the parents that must be included in the product. Since these functions do not represent conditional distributions, they are characterized by having only incoming edges. For example, in Fig. 3b the function $n(x, y)$ is a function used to normalize a conditional distribution for $z$. The fact that $n(x, y)$ is used to normalize a conditional distribution on $z$ is indicated by a dashed edge connecting $n(x, y)$ to $z$. Generally, if a variable has two or more incoming edges that are connected to functions that each have exactly one outgoing edge, the product of the functions and any associated normalization functions connected to the parents, must be normalized w.r.t. the variable. In Fig. 3b, we require that $f(x, z)g(y, z)n(x, y)$ be normalized w.r.t. $z$.

The above situations correspond to single functions and multiple children or multiple functions with a single child. The general normalization condition applies to situations where there are sets of functions and corresponding sets of children. Generally, in a directed factor graph, every maximal set of functions and variables that are related by a connected graph with directed edges from functions to variables, must satisfy the condition that the product of the functions and the associated normalization functions are normalized w.r.t. the set of variables. This condition can be viewed as a factor graph equivalent of the chain graph normalization (Lauritzen 1996).

### 3.2 Converting Factor Graphs to BNs

If every variable in a directed factor graph has exactly one incoming directed edge, the factor graph can be converted to a BN in a straightforward fashion, as described previously. If a variable has more than one neighboring function that is connected by a directed edge, the conditional probability of the child given its parents is set to the product of these functions and any associated normalization functions (which may be indicted by dashed edges). Since a directed factor graph cannot have directed cycles, this process ensures that the resulting BN does not have directed cycles and is properly normalized.

### 3.3 Converting Factor Graphs to MRFs

Any factor graph can be converted to an MRF. For each function node, a clique is created on the neighboring variables. Then, the maximal cliques are identified and each function is associated with one of the maximal cliques that includes the variables neighboring the function node. The potential for each maximal clique is set to the product of the functions that are associated with the maximal clique. This procedure works for undirected factor graphs, directed factor graphs, and hybrid factor graphs. For example, this procedure can be used to convert the factor graphs in Fig. 2b, Fig. 2c, Fig. 2d, and Fig. 2f to the MRF in Fig. 2e.

Sometimes, a choice is made as to which maximal clique a function should be associated with. While there are no such choices when converting the factor graphs in Fig. 2d and Fig. 2f, for the factor graphs in Fig. 2b and Fig. 2c, we can either set $\phi(u, v) = P(u)P(v|u)$ and $\phi(u, x) = P(x|u)$ or set $\phi(u, v) = P(v|u)$ and $\phi(u, x) = P(u)P(x|u)$. In fact, we can also choose to associate portions of functions with maximal cliques, e.g., set $\phi(u, v) = P(u)^{1/2}P(v|u)$ and $\phi(u, x) = P(u)^{1/2}P(x|u)$.

### 3.4 More Specificity About Factorization

Since factor graphs have function nodes, each of which corresponds to a term in the factorization of the joint distribution, factor graphs can express arbitrary factorizations of the joint distribution. This is not true for BNs, MRFs and chain graphs. It is easy to construct interesting and useful factor graphs that, when converted to BNs or MRFs, are not as expressive about factorization. For example, if the factor graph in Fig. 5c is converted to a BN as described in a previous section, the BN in Fig. 5a is obtained. This BN does not express the fact that $P(z|m, c_1, c_0)$ factorizes into a function of $m$, $c_1$ and $z$ and another function of

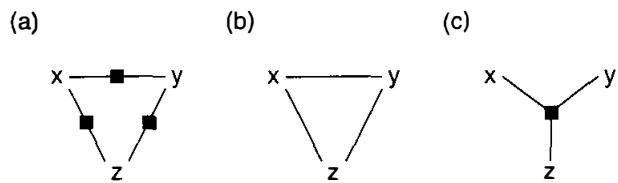

Figure 4: Factor graphs are more explicit about factorization than are MRFs. (a) A factor graph that indicates the factorization $f(x,y)g(y,z)h(x,z)$. (b) The MRF obtained by creating a clique for each function node in the factor graph (here, the cliques are pairs of variables). Unfortunately, these cliques are not maximal, so the resulting MRF cannot identify the correct factorization. (c) The factor graph obtained by identifying the maximal cliques in the MRF.

$m$, $c_0$ and $z$.

As another example, consider the factor graph in Fig. 4a, which describes the factorization $P(x,y,z) = f(x,y)g(y,z)h(x,z)$. The corresponding MRF is shown in Fig. 4b. However, this MRF has one maximal clique, which fails to express the more detailed factorization of the joint distribution, as shown by converting the MRF back to a factor graph (Fig. 4c).

In terms of factorization, chain graphs consist of components that express factorization in a similar manner as both BNs and MRFs. As with the latter models, chain graphs cannot express arbitrary factorizations.

Several important theoretical and practical problems are characterized by joint distributions that are given by the product of a very large number of functions, including error-correcting decoding (MacKay and Neal 1996; Frey and Kschischang 1996; McEliece 1997; Frey and MacKay 1998), phase unwrapping (Frey, Koetter and Petrovic 2002), and satisfiability (Mézard, Parisi and Zecchina 2002).

While extra dummy variables can be added to BNs and MRFs to assist them in modeling arbitrary factorizations, the use of dummy variables is clumsy (requiring additional, nonuniversal notation). Importantly, introducing these dummy variables also obscures conditional independence relationships.

## 4 ASCERTAINING CONDITIONAL INDEPENDENCE

The rule for ascertaining conditional independence are obtained by judiciously merging the rules for BNs and MRFs. The resulting rule works for directed factor graphs, undirected factor graphs, and partially-directed factor graphs. In this section, after describing the rule, we give some examples, including a commonly used model component whose conditional independencies cannot be completely represented in a BN or an MRF, but can be represented in a factor graph. We also describe how a factor graph can represent the conditional independencies in a chain graph, while revealing more detail of the factorization of the distribution.

To determine whether two variables are independent given a set of observed variables (the variables that are conditioned on), consider all paths connecting the two variables. If all paths are blocked, the variables are conditionally independent, and otherwise, they may not be conditionally independent. A path is blocked if any one or more of the following conditions are satisfied:

1. One of the variables in the path is observed

2. One of the variables or functions in the path has two incoming edges that are part of the path, and neither the variable or function nor any of its descendants are observed

It is interesting that the set of 3 conditions for blockage in BNs (Pearl 1988) reduces to a set of 2 conditions in factor graphs, and the first condition covers the condition for blockage in MRFs.

For example, from the factor graph in Fig. 2c, we see that $x$ and $y$ are independent given $u$ and $v$. There are two paths from $x$ to $y$ and the upper path is blocked by the observation of $u$ (Cond 1). The lower path is blocked by Cond 2, because there is a function node with 2 incoming directed edges in the path and the descendants of the function (namely $z$) are not observed. (The function itself is also not "observed".)

The factor graph in Fig. 2f was derived from the MRF for the joint distribution (Fig. 2e) and gives a different answer to the above problem. In this case, the upper path is blocked, but the lower path is not. None of the variables along the lower path are observed, and there are no variables or functions with incoming directed edges along the path, so Cond 2 is not applicable.

In the above example, the directed factor graph in Fig. 2c indicates a conditional independence that is lost in the factor graph in Fig. 2f. We now make some general observations about ascertaining independencies in factor graphs.

### 4.1 Unification of Conditional Independence Relationships

Every BN has a corresponding directed factor graph that can be converted back to the BN. It is straightforward to show that in this case, the above rules for ascertaining conditional independence reduce to the rules for BNs. It follows that the factor graph rules can be used to ascertain any conditional independencies that can be ascertained in a BN. The same argument holds for MRFs. So, the union of the sets of models whose independencies can be expressed by BNs or MRFs is a subset of the models whose independencies can be expressed by factor graphs. In this sense, factor graphs subsume BNs and MRFs.

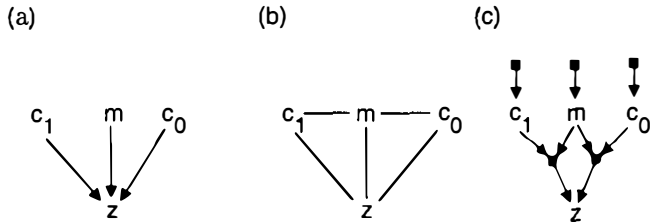

Figure 5: (a) A BN for the mixture-of-experts model. (b) A MRF for the mixture-of-experts model. (c) A factor graph for the mixture-of-experts model.

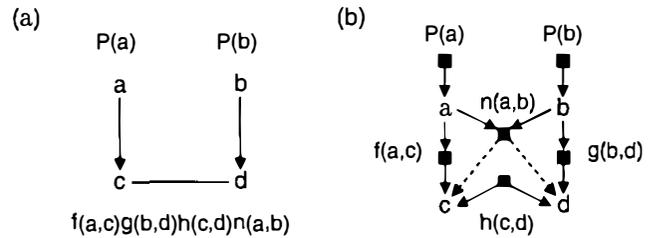

Figure 6: (a) A chain graph from (Lauritzen 1996). (b) A factor graph that indicates all of the conditional independencies that are indicated by the chain graph, but is more specific about the factorization of the joint distribution.

In fact, it turns out there BNs and MRFs form a strict subset of factor graphs: There are models whose conditional independencies cannot be completely expressed in BNs or MRFs, but can be expressed in factor graphs. A common model component consists of a "switch" that decides which variable is the parent of another variable. For example, this component is found in the mixture-of-experts model (Jacobs et al. 1991). Suppose $z$ is the child, $m$ is the switch, and $c_1$ and $c_0$ are the two potential parents. If $m = 0$, the parent of $z$ is $c_0$, and if $m = 1$, the parent of $z$ is $c_1$. Assuming $m$, $c_0$ and $c_1$ are independent, the joint distribution can be written

$$P(z, m, c_0, c_1) = P(z|c_1)^m P(z|c_0)^{1-m} P(m) P(c_0) P(c_1).$$

Note that for $m = 1$, we get $P(z, m, c_0, c_1) = P(z|c_1) P(m) P(c_0) P(c_1)$ and for $m = 1$, we get $P(z, m, c_0, c_1) = P(z|c_0) P(m) P(c_0) P(c_1)$.

In this model, $m$, $c_0$ and $c_1$ are independent. Also, $c_1$ and $c_0$ are independent given $m$, since if the state of the switch is known, $z$ is related to one and only one of the two variables $c_0$ and $c_1$ so they are uncoupled. Also, $c_1$ and $c_0$ are independent given $m$ and $z$.

Fig. 5a and Fig. 5b show the BN and the MRF for this model. The BN clearly indicates that $m$, $c_0$ and $c_1$ are marginally independent, but fails to indicate that given $z$ and $m$, $c_0$ and $c_1$ are independent. According to the rule for BNs, the path from $c_1$ to $z$ to $c_0$ is not blocked, when $z$ is observed. The MRF clearly indicates that $c_1$ and $c_0$ are independent given $m$ and $z$, but fails to indicate that $c_1$, $m$ and $c_0$ are independent. Further, like the BN, the MRF fails to indicate that $c_1$ and $c_0$ are independent given $m$.

A factor graph for this problem is shown in Fig. 5c. The singly-connected function nodes correspond to the marginals $P(c_1)$, $P(m)$ and $P(c_0)$. The other two functions are $P(z|c_1)^m$ and $P(z|c_0)^{1-m}$. According to the normalization condition for directed factor graphs, we require that the product of these two functions be normalized with respect to $z$. This is true, since $\sum_z P(z|c_1)^m P(z|c_0)^{1-m} = 1$.

It turns out that this factor graph expresses *all* of the conditional independencies for the above distribution. Without any observations, the paths between $c_1$, $m$ and $c_0$ are blocked by Cond 2, since they either include a function node with two incoming directed edges and unobserved descendants $z$, or they include a variable node with two incoming directed edges, where neither the variable ($z$) nor any of its descendants (none) are observed. $c_1$ and $c_0$ are also independent when $m$ is observed, since the upper path that goes through $m$ is still blocked (in fact, blocked "more" by the observation of $m$ in Cond 1). When both $m$ and $z$ are observed, $c_1$ and $c_0$ are independent. The upper path from $c_0$ to $c_1$ is blocked by the observation of $m$ in Cond 1, and the lower path is blocked by $z$ in Cond 1. When only $z$ is observed, $c_1$ and $c_0$ are not independent. The upper path is no longer blocked by $m$ in Cond 1, and although there are function nodes with two incoming edges, in both cases there is a descendant, $z$, that is observed, preventing Cond 2 from being satisfied.

This example shows that the set of models whose independencies can be expressed by factor graphs is a strict superset of the union of such models for BNs and MRFs.

### 4.2 Expressing Chain Graph Independencies

Chain graphs can represent conditional independencies that cannot be represented by BNs or MRFs. Fig. 6a shows a chain graph that is frequently used to demonstrate this representational power (Lauritzen 1996). This chain graph describes the distribution $P(a, b, c, d) = P(a) P(b) P(c, d|a, b)$ with $P(c, d|a, b) = f(a, c) g(b, d) h(c, d) n(a, b)$, where $n(a, b)$ is a normalizing function: $n(a, b) = 1/\sum_c \sum_d f(a, c) g(b, d) h(c, d)$. In this model, $a$ and $b$ are independent, $a$ and $d$ are independent given $b$ and $c$, and $b$ and $c$ are independent given $a$ and $d$. These independencies can be de-termined using the conditional independence rule for chain graphs (Lauritzen 1996).

While the above chain graph efficiently describes a set of conditional independencies, it does not accurately reflect the factorization of the distribution, which consists of 6 terms. The factor graph in Fig. 6b explicitly represents the 6 terms using 6 function nodes. Note the dashed edges, which indicate that $n(a,b)$ is used to normalize the conditional distribution over $c, d$.

This factor graph indicates the same independencies as the chain graph. $a$ and $b$ are independent, since both of the two paths between $a$ and $b$ are blocked. The lower path is blocked by Cond 2 because c has two incoming edges and is not observed. The upper path is blocked by Cond 2 because $n(a,b)$ has two incoming edges and none of its descendants (c and $d$) are observed. $a$ and $d$ are independent given $b$ and c, since both paths between $a$ and $d$ are blocked. The lower path is blocked by Cond 1 because c is observed. The upper path is blocked by Cond 1 because $b$ is observed. Similarly, $b$ and c are independent given $a$ and $d$. Note that $a$ and $b$ are not independent given c or $d$ or both, because the upper path from $a$ to $b$ is not blocked whenever c or $d$ or both c and $d$ are observed, by Cond 2.

## 5 CONCLUSIONS

We have shown that an extension to factor graphs is able to unify the properties Bayesian networks (BNs) and Markov random fields (MRFs), and is more specific about factorization of the joint distribution than either BNs or MRFs. A factor graph can represent any set of conditional independencies that can be represented by a BN or an MRF, and there exist important models whose independencies cannot be represented by BNs or MRFs, but can be represented by a factor graph. We show how a factor graph can also be used to represent the conditional independencies in a chain graph. An advantage of factor graphs, is that they can be constructed using causal knowledge or non-causal knowledge or both. We emphasize that factor graphs are not meant to be a replacement for BNs, MRFs or chain graphs (the author draws BNs and MRFs all the time). However, there are many circumstances where BNs, MRFs and chain graphs are inadequate for expressing graphical structure, but where factor graphs excel.

## ACKNOWLEDGMENTS

We thank Max Chickering, David Heckerman, Michael Jordan, Frank Kschischang, Chris Meek, Radford Neal, and Max Welling for helpful discussions. We also thank an anonymous reviewer for urging us to investigate how factor graphs compare to chain graphs, and specifically the chain graph $a \to c - d \leftarrow b$.

## References

Dechter, R. 1996. Bucket elimination: A unifying framework for probabilistic inference. In *Proceedings of the Twelvth Conference on Uncertainty in Artificial Intelligence*. Morgan Kaufmann.

Frey, B. J., Koetter, R., and Petrovic, N. 2002. Very loopy belief propagation for unwrapping phase images. In *2001 Conference on Advances in Neural Information Processing Systems, Volume 14*. MIT Press.

Frey, B. J. and Kschischang, F. R. 1996. Probability propagation and iterative decoding. In *Proceedings of the 1996 Allerton Conference on Communication, Control and Computing*.

Frey, B. J., Kschischang, F. R., Loeliger, H. A., and Wiberg, N. 1998. Factor graphs and algorithms. In *Proceedings of the $35^{th}$ Allerton Conference on Communication, Control and Computing 1997*.

Frey, B. J. and MacKay, D. J. C. 1998. Trellis-constrained codes. In *Proceedings of the $35^{th}$ Allerton Conference on Communication, Control and Computing 1997*.

Gallager, R. G. 1963. *Low-Density Parity-Check Codes*. MIT Press, Cambridge MA.

Jacobs, R. A., Jordan, M. I., Nowlan, S. J., and Hinton, G. E. 1991. Adaptive mixtures of local experts. *Neural Computation*, 3:79–87.

Kinderman, R. and Snell, J. L. 1980. *Markov Random Fields and Their Applications*. American Mathematical Society, Providence USA.

Kschischang, F. R., Frey, B. J., and Loeliger, H.-A. 2001. Factor graphs and the sum-product algorithm. *IEEE Transactions on Information Theory, Special Issue on Codes on Graphs and Iterative Algorithms*, 47(2):498–519.

Lauritzen, S. L. 1996. *Graphical Models*. Oxford University Press, New York NY.

MacKay, D. J. C. and Neal, R. M. 1996. Near Shannon limit performance of low density parity check codes. *Electronics Letters*, 32(18):1645–1646. Reprinted in *Electronics Letters*, vol. 33, March 1997, 457–458.

McEliece, R. J. 1997. Coding theory and probability propagation in loopy Bayesian networks. Invited talk at *Thirteenth Conference on Uncertainty in Artificial Intelligence*.

Mézard, M., Parisi, G., and Zecchina, R. 2002. Analytic and algorithmic solution of random satisfiability problems. *Science*, 297:812–815.

Pearl, J. 1988. *Probabilistic Reasoning in Intelligent Systems*. Morgan Kaufmann, San Mateo CA.

Yedidia, J., Freeman, W. T., and Weiss, Y. 2001. Generalized belief propagation. In *Advances in Neural Information Processing Systems 13*. MIT Press, Cambridge MA.